%% file: main.tex
\documentclass[10pt,twocolumn,letterpaper]{article}
\usepackage[pagenumbers]{cvpr}
\input{preamble}
\definecolor{cvprblue}{rgb}{0.21,0.49,0.74}
\usepackage[breaklinks,colorlinks,allcolors=cvprblue]{hyperref}

\title{TalentRL: Strategy-Guided Reinforcement Learning\\
for Hierarchical Skill Routing}

\author{Huancheng Chen, Xiaodi Sun, Zhaoqiong Huang, Michelle Chang, Jingwen Lu\\
Microsoft AI\\
{\tt\small huancchen@microsoft.com}
}

\hypersetup{
  pdftitle={TalentRL: Strategy-Guided Reinforcement Learning for Hierarchical Skill Routing},
  pdfauthor={Huancheng Chen, Xiaodi Sun, Zhaoqiong Huang, Michelle Chang, Jingwen Lu}
}

\begin{document}

\maketitle

\input{src/abstract}
\input{src/introduction}
\input{src/related_work}
\input{src/method}
\input{src/reinforcement_learning}
\input{src/experiment}
\input{src/conclusion}

{
\small
\bibliographystyle{ieeenat_fullname}
\bibliography{main}
}

\onecolumn
\appendix

\section{Experimental details}
\label{app:experimental-configuration}

\begin{table}[ht]
\caption{Primary Qwen2.5-7B Search7 training, routing, and retrieval configuration.
Interactive-task protocols are described separately below.}
\label{tab:training-config}
\centering
\small
\begin{tabular}{llr}
\toprule
Stage & Hyperparameter & Value \\
\midrule
SFT & Epochs & 2 \\
SFT & Learning rate & $1\times10^{-5}$ \\
SFT & Effective batch size & 16 \\
SFT & Maximum sequence length & 8,192 \\
\midrule
RL & Policy updates & 200 \\
RL & Learning rate & $1\times10^{-6}$ \\
RL & Task batch size & 64 \\
RL & Rollouts per task & 8 \\
RL & KL coefficient & 0.01 \\
RL & Policy-ratio clipping radius & 0.2 \\
RL & Maximum interaction steps & 4 \\
\midrule
Routing & Reduced skill budget $k$ & 2 \\
Routing & Dead-zone threshold $\eta$ & 0.1 \\
Routing & Advantage scale $\alpha$ & 2.0 \\
Routing & Advantage clipping bound $\kappa$ & 2.0 \\
\midrule
Retrieval & Embedding model & Qwen3-Embedding-0.6B \\
Retrieval & Similarity recall size & 20 \\
Retrieval & Execution skill budget & 5 \\
Retrieval & Similarity threshold & 0.2 \\
Retrieval & Talent-prior weight $\lambda$ & 0.3 \\
\bottomrule
\end{tabular}
\end{table}

Training rollouts are sampled to produce within-task diversity for
group-relative optimization; groups in which all rollouts receive the same
return receive zero execution advantage. Final Search7 evaluation uses
deterministic decoding. Skill and query embeddings are L2-normalized, so
$q(x,s)$ is cosine similarity. The three routing contrasts are equally
weighted. The full generated talent response receives the routing advantage.

\paragraph{Search7 baseline configurations.}
Table~\ref{tab:rl-summary} uses our Search7 adaptations of the baselines,
aligned with the official implementations. As described in
Section~\ref{sec:main-results}, TalentRL starts from TalentRL-SFT, whereas
the baselines start from the corresponding SkillRL-SFT checkpoints.
TalentRL is evaluated after 200 RL steps.

\paragraph{Interactive-task baseline results.}
Baseline values in Table~\ref{tab:interactive-results} are reported by
SkillRL \citep{xia2026skillrl}, except the Skill1 row, which
comes from \citet{shi2026skill1}.
Skill1 uses Qwen2.5-7B-Instruct and averages results over three runs.
These published results are not protocol-matched reproductions.
Highlighting includes all rows:
the highest and second-highest distinct values for each metric are
bold and underlined, respectively, with ties receiving the same formatting.

\paragraph{Interactive-task evaluation protocols.}
Plain SFT and TalentRL-SFT use Qwen2.5-7B-Instruct without interactive RL.
TalentRL continues each TalentRL-SFT checkpoint with interactive GRPO and
the evolving SkillBank for 150 steps on each benchmark; we evaluate the
step-150 checkpoints with the SkillBank state saved at that step and the
same evaluation protocol as TalentRL-SFT.
Plain SFT fine-tunes all parameters on successful GPT-5.5 trajectories
with embedding-retrieved skills (batch 16, two epochs, learning rate
$10^{-5}$). TalentRL-SFT continues this checkpoint with the same recipe on
talent-labeled rows only: 4,961 ALFWorld and 1,212 WebShop rows, each task
contributing one routing example and its talent-conditioned actions.
The supervision comes from matched rollouts of the Plain SFT student
under no talent and under each of the five ALFWorld or five WebShop talents,
with three repeats per condition; each task keeps the talent with the
highest positive mean utility, breaking ties by score and then by
fewer turns. For ALFWorld, the training tasks are resampled so that each
of the five talents labels 85--86 of the 429 training tasks.
During TalentRL-SFT and TalentRL evaluation, the model first selects a
talent, then acts with its guidance and five skills reranked from the 20 most task-similar skills
with a bonus $\lambda=0.3$ for the talent's skills; an invalid route falls back to
no talent. Talent selection uses greedy decoding; execution responses use
temperature 0.4 and up to 512 tokens per response.
Plain SFT uses embedding-retrieved skills without talent routing or guidance.
On ALFWorld, the table reports 140 held-out \texttt{valid\_seen} tasks with
state-feedback prompts and a 50-step limit; on the 134
\texttt{valid\_unseen} tasks, Plain SFT succeeds on 131 and TalentRL-SFT on 133.
On WebShop, evaluation covers the first 500 goals in the canonical seed-42
ordering within the 1,000-product environment with a 15-step limit.
All tasks remain in the denominator. TalentRL-SFT succeeds on 131/140
ALFWorld and 315/500 WebShop tasks, with mean WebShop reward 0.7582
(75.8 on the table's 0--100 scale); all TalentRL-SFT routes are valid.
After interactive RL, TalentRL succeeds on 139/140 ALFWorld games (99.3\%)
and 398/500 WebShop goals (79.6\%), with a WebShop average score of 90.7,
as reported in Table~\ref{tab:interactive-results}.
These are single-checkpoint native-protocol evaluations, not
protocol-matched reproductions of the published baselines.

\section{Talent taxonomy}
\label{app:taxonomy}
The initial Search7 catalog uses eight talents:
\talentname{task decomposition}, \talentname{bridge-entity resolution},
\talentname{entity disambiguation}, \talentname{constraint application},
\talentname{evidence comparison}, \talentname{long-tail query expansion},
\talentname{temporal and numeric reasoning}, and
\talentname{answer verification}. Each talent contains a
natural-language description, generation guidance, and references to a
subset of reusable skills.

\subsection{Preset talent-to-skill assignments}
Table~\ref{tab:preset-talent-skills} reports the talent-specific generation
guidance and initial talent--skill associations. The guidance conditions
execution, while the associations supply a soft
prior rather than a hard retrieval boundary. The retriever recalls the 20
most similar skills above a similarity threshold of 0.2, adds the bonus
$\lambda=0.3$ to skills associated with the policy-selected talent, and
supplies the top five reranked skills during execution. Skills, talents, and associations are updated during
the reported RL runs, starting from these preset assignments.

\begin{table*}[t]
\caption{Talent-specific generation guidance and preset talent--skill
associations for Search7. The selected talent provides natural-language
guidance for execution and a soft prior for skill retrieval. Associated
skills are not necessarily all included in the execution context.
Table~\ref{tab:skill-card-details} provides the referenced skill details.}
\label{tab:preset-talent-skills}
\vspace{6pt}
\centering
\small
\renewcommand{\arraystretch}{1.2}
\setlength{\tabcolsep}{4pt}
\setlength{\defaultaddspace}{0.65em}
\begin{tabular}{@{}
>{\raggedright\arraybackslash}p{0.24\textwidth}
>{\raggedright\arraybackslash}p{0.50\textwidth}
>{\raggedright\arraybackslash}p{\dimexpr0.26\textwidth-4\tabcolsep\relax}@{}}
\toprule
\textbf{Talent} & \textbf{Generation guidance} & \textbf{Associated skills} \\
\midrule
\talentname{Task decomposition}
&
Identify the dependency chain before searching, solve one subproblem at a
time, and carry forward only verified intermediate facts.
&
\texttt{gen\_001}, \texttt{mul\_001}, \texttt{mul\_002},
\texttt{gen\_007}, \texttt{gen\_010} \\
\addlinespace
\talentname{Bridge-entity resolution}
&
Resolve the bridge entity first, preserve relation direction, then issue a
targeted follow-up query for the requested attribute.
&
\texttt{gen\_007}, \texttt{mul\_001}, \texttt{mul\_002},
\texttt{gen\_004}, \texttt{gen\_010} \\
\addlinespace
\talentname{Entity disambiguation}
&
Use type, date, location, occupation, or related-work qualifiers and reject
evidence that belongs to another referent.
&
\texttt{gen\_006}, \texttt{mul\_004}, \texttt{ent\_002},
\texttt{ent\_003}, \texttt{gen\_005} \\
\addlinespace
\talentname{Constraint application}
&
Translate every clause into a checkable constraint, narrow candidates by
intersection, and verify all constraints before answering.
&
\texttt{com\_004}, \texttt{mul\_001}, \texttt{mul\_002},
\texttt{gen\_005}, \texttt{gen\_010} \\
\addlinespace
\talentname{Evidence comparison}
&
Collect the same attribute for every entity, normalize dates or units, and
perform the comparison only after all values are verified.
&
\texttt{gen\_008}, \texttt{mul\_003}, \texttt{com\_001},
\texttt{com\_002}, \texttt{com\_003}, \texttt{com\_005},
\texttt{gen\_004} \\
\addlinespace
\talentname{Long-tail query expansion}
&
When direct retrieval is weak, try aliases, related works, distinctive
phrases, and progressively refined contextual queries.
&
\texttt{dir\_002}, \texttt{dir\_003}, \texttt{ent\_004},
\texttt{gen\_003}, \texttt{mul\_005}, \texttt{gen\_006} \\
\addlinespace
\talentname{Temporal and numeric reasoning}
&
Retrieve authoritative dated or numeric evidence, normalize formats and
units, and explicitly compute or compare the required values.
&
\texttt{gen\_009}, \texttt{dir\_004}, \texttt{gen\_008},
\texttt{com\_003}, \texttt{com\_005}, \texttt{gen\_005} \\
\addlinespace
\talentname{Answer verification}
&
Do not guess from memory or a weak snippet; obtain explicit supporting
evidence, cross-check critical facts, and stop once the answer is solid.
&
\texttt{gen\_004}, \texttt{gen\_005}, \texttt{gen\_010},
\texttt{dir\_005}, \texttt{ent\_003} \\
\bottomrule
\end{tabular}
\end{table*}

\clearpage
\subsection{Skill details}
Table~\ref{tab:skill-card-details} lists the principles and applicability
conditions of the 26 distinct skills referenced in
Table~\ref{tab:preset-talent-skills}. Each skill is listed once, even when it
is associated with multiple talents.
\input{src/skill_card_details}

\subsection{Talent selection distribution}
\label{app:talent-distribution}
Figure~\ref{fig:talent-distribution} reports the talents selected at
evaluation by the Qwen3-8B TalentRL policy on Search7 and by the final
TalentRL policies on ALFWorld and WebShop. All selections use greedy
decoding. The Search7 distribution is discussed in
Section~\ref{sec:main-results}; this subsection analyzes the interactive tasks.

\paragraph{ALFWorld.}
Panel~(b) groups ALFWorld tasks by type over the seen (140) and unseen (134)
games. The selected talent follows the step most likely to fail in each
task type. Pick and Pick2 tasks only require finding objects and placing
them, so their main difficulty is searching drawers, cabinets, and
countertops; they use \talentname{systematic receptacle search} in 83\% and 100\% of
games, respectively, and Pick2 additionally requires locating a second
instance of the same object type. Clean tasks are routed mainly to
\talentname{exact object selection} (71\%). Cleaning itself takes a single action at the sink
basin, and many Clean targets are kitchen utensils with similar names, such
as knife and butter knife or spoon and ladle, so selecting the correct
object matters more than the cleaning step. Heat tasks require operating the
microwave, the most involved appliance interaction among the six task types,
and select \talentname{fixture interaction} more often than any other task
type (23\%). Cool tasks use the fridge, which needs simpler interaction, and
are routed mainly to \talentname{systematic receptacle search} (61\%). Look
tasks require finding an object and then turning on a desk lamp, and are split
between \talentname{systematic receptacle search} (45\%) and
\talentname{exact object selection} (48\%). \talentname{Goal decomposition}
and \talentname{destination delivery} are not selected: ALFWorld tasks follow short fixed templates that need no further
decomposition, and the final placement step rarely fails. Because TalentRL
already succeeds on 139 of the 140 seen games, these distributions show the
learned routing preference rather than evidence that another talent would
fail on the same task.

\paragraph{WebShop.}
WebShop talents correspond to product domains, so the selection follows the
category of the requested product, and the distribution mainly reflects the
composition of the 500 test goals. All 322 goals routed to \talentname{apparel} request
clothing items, such as shirts, hoodies, dresses, and activewear. \talentname{Home decor}
receives furniture, window coverings, and bedding (76 goals), and \talentname{footwear}
receives shoes, boots, and sandals (65 goals). The 31 goals routed to \talentname{generic
matching} fall outside the four specific domains: about 17 request
electronics, such as cables, phone accessories, and storage devices, about 9
request food, and the rest are mostly beauty products. The \talentname{personal care}
talent is defined narrowly around personal-care items and travel
containers, and is selected for only 6 skincare, hair-removal, and
hair-growth goals; similar beauty goals are sometimes routed to
\talentname{generic matching} instead. Success rates range from 61.5\% for footwear to 83.9\%
for apparel. Footwear goals are the hardest because they require selecting
exact size and color variants, and a single wrong option prevents a full
match.

\section{Search7 SFT data construction}
\label{app:sft}
This section details the Search7 construction in Section~\ref{sec:sft};
the interactive-task SFT stages are described in
Appendix~\ref{app:experimental-configuration}.
We start from the SkillRL Search SFT split, group its step-level rows by
question, and reconstruct task trajectories. For each task $x$, we collect
$n$ rollouts for every forced talent $c$ and $n$ reference rollouts with
similarity-retrieved skills but no talent guidance or retrieval bonus.
Following Section~\ref{sec:sft}, we compute
$u_{c,r}(x)=R(\tau_{c,r})-R(\tau_{\varnothing,r})$ and average over repeats.
We select the talent with the largest mean utility and retain the task
only if this utility is positive, breaking talent ties by mean conditioned score,
positive-outcome rate, fewer tool calls, and a deterministic task--talent
hash. We then choose one normally completed, valid, positive-utility
trajectory for that talent, prioritizing higher utility and fewer tool calls.
Normal completion and action validity are eligibility requirements, not
tie-breaking preferences.

Each accepted paired trajectory yields a talent-selection row and one
ReAct row for every search or answer action. The routing target contains a
task-grounded rationale in \texttt{<think>} and the selected talent identifier
in \texttt{<talent>}. Skills are
selected by the same talent-conditioned similarity retriever used at
inference and therefore require no policy target. The ReAct targets preserve
the accepted talent-conditioned rollout rather than the original SkillRL
actions. Tasks without a positive talent do not produce routing-aware rows.
Original SkillRL rows for the source tasks are retained as replay, and
the two sources are concatenated without enforcing a fixed ratio.

Training and validation are separated by a deterministic task hash, ensuring
that no task appears in both partitions. The final Search7 evaluation sets are
not used to construct routing targets. We initialize from the base instruction
model and run two epochs of full-parameter, response-only SFT with a low
learning rate. Prompt tokens and injected skill context are excluded from the
loss. Replay rows retain the original search protocol without adding a
talent-selection target.

\section{Matched-intervention RL details}
\label{app:rl-data}
The RL source row contains the task prompt and benchmark metadata. For each
task, the system creates an anchor rollout, matched no-skill,
reduced-context, and no-prior controls, and ordinary policy rollouts.
Retrieved observations, talent guidance, and injected skills are environment
context and are masked out of the policy loss.

\subsection{Anchor and control rollouts}
The anchor uses the policy-selected talent, its guidance, and five
talent-conditioned skills. The no-skill control retains the talent
and guidance but withholds skills. The reduced-context control ($k=2$)
retains the same talent, guidance, and retrieval ordering while supplying
only the top two skills. The no-prior control removes talent guidance and
the retrieval bonus, supplying five skills retrieved by similarity alone.
The matched controls are retained for routing-credit estimation, but all
their response tokens have zero surrogate advantage, as specified by
Equation~\ref{eq:token-advantage}. Only the anchor's routing response,
including both its rationale and talent identifier, receives the routing
advantage; ordinary rollouts' routing tokens have zero surrogate advantage.
Execution tokens in the anchor and ordinary rollouts receive the
group-relative advantage computed from those rollouts only, excluding
the matched controls.

\subsection{Reward and credit details}
\label{app:reward}
The return is the task score alone, with no additional reward shaping.
Talent routing credit averages three signed, dead-zoned return differences,
namely the anchor return minus the no-skill, reduced-context, and no-prior
control returns. Following Equation~\ref{eq:routing-advantage}, we multiply
this average by $\alpha=2.0$, then clip it to $[-\kappa,\kappa]$ with
$\kappa=2.0$ before applying it to the anchor routing response.
We use $\eta=0.1$ and equal contrast weights.
The shared policy is optimized with the masked surrogate and KL
regularization in Equation~\ref{eq:rl-loss}; the reference policy is
initialized from the SFT checkpoint and frozen throughout RL.
The intervention contrasts are auxiliary routing-utility signals rather
than unbiased estimators of the original episodic-return advantage.

\end{document}

%% file: preamble.tex
\usepackage[T1]{fontenc}
\input{math_commands.tex}

\usepackage{algorithm}
\usepackage{algpseudocode}
\usepackage{array}
\usepackage{microtype}
\usepackage{multirow}
\usepackage{longtable}
\usepackage[most]{tcolorbox}

\definecolor{routingbg}{RGB}{242,246,255}
\definecolor{routingframe}{RGB}{84,112,168}
\definecolor{thinktag}{RGB}{43,103,176}
\definecolor{talenttag}{RGB}{38,135,90}
\definecolor{skilltag}{RGB}{38,135,90}
\definecolor{creditbg}{RGB}{255,247,232}
\definecolor{creditframe}{RGB}{183,121,31}
\definecolor{updatebg}{RGB}{238,249,241}
\definecolor{updateframe}{RGB}{59,132,82}
\newtcolorbox{routingresponse}[1]{
  enhanced,
  colback=routingbg,
  colframe=routingframe,
  colbacktitle=routingframe,
  coltitle=white,
  boxrule=0.5pt,
  arc=1.2mm,
  boxsep=0mm,
  left=2mm,
  right=2mm,
  top=1.1mm,
  bottom=1.1mm,
  toptitle=0.7mm,
  bottomtitle=0.7mm,
  titlerule=0pt,
  before skip=0.65em,
  after skip=0.65em,
  fonttitle=\bfseries\small,
  before upper={\raggedright\setlength{\parindent}{0pt}\setlength{\parskip}{0.15em}},
  title={#1}
}
\newcommand{\routingthinktag}[1]{%
  \textcolor{thinktag}{\texttt{\textless#1\textgreater}}}
\newcommand{\routingtalenttag}[1]{%
  \textcolor{talenttag}{\texttt{\textless#1\textgreater}}}

\newcommand{\talentname}[1]{\textsc{#1}}

%% file: math_commands.tex
\usepackage{amsmath,amsfonts,bm}

\def\eqref#1{equation~\ref{#1}}

\def\1{\bm{1}}

\DeclareMathAlphabet{\mathsfit}{\encodingdefault}{\sfdefault}{m}{sl}
\SetMathAlphabet{\mathsfit}{bold}{\encodingdefault}{\sfdefault}{bx}{n}



%% file: src/abstract.tex
\begin{abstract}
Natural-language skills enable large language model (LLM) agents to reuse
procedures across tasks, yet most systems retrieve individual skills from a
flat SkillBank without explicitly modeling the high-level strategy that
governs their selection and use. We introduce TalentRL, a strategy-guided
reinforcement learning framework for hierarchical routing over procedural
memory. Coarse-grained \emph{talents} represent recurring strategies for skill
use, while fine-grained \emph{skills} provide executable procedures. For
each task, the policy model selects a talent that guides execution and softly
prioritizes its associated skills during similarity-based retrieval. However,
since task outcomes depend jointly on talent selection, retrieved skill
context, and downstream execution, terminal rewards alone cannot isolate the
contribution of the routing decision. TalentRL therefore trains talent routing
with reward contrasts between rollouts of the same task under matched
interventions on the talent-conditioned context, and optimizes downstream
execution with group-relative advantages. Across seven open-domain
and multi-hop question-answering benchmarks, talent-aware supervised
fine-tuning improves average success rate by 7.53\%, 6.71\%, and 9.21\%
over the corresponding zero-shot baselines for Qwen2.5-7B, Qwen3-8B,
and Llama-3.1-8B, respectively. Reinforcement learning further improves these
SFT checkpoints by 7.58\% to 9.32\%, and TalentRL achieves the
highest average success rate among all compared methods on all three
backbones. On the agentic
benchmarks ALFWorld and WebShop, TalentRL achieves the highest ALFWorld
success rate (99.3\%) and the highest WebShop score (90.7) among all compared
methods.
\end{abstract}

%% file: src/introduction.tex
\section{Introduction}

Large language model (LLM) agents solve interactive tasks by interleaving
reasoning, retrieval, and environment actions \citep{yao2023react}. To improve
behavior across episodes, verbal-memory methods retain linguistic reflections
from prior interactions \citep{shinn2023reflexion}, while outcome-based
reinforcement learning (RL) optimizes the policy using task rewards
\citep{jin2025searchr1}. Procedural-memory systems further store reusable
skills distilled from past experience, yet most retrieve individual skills from a flat SkillBank without having the policy model
explicitly select a high-level strategy to guide their selection and use.
This motivates a central question: how can an agent learn, from task-level
outcomes, to select a high-level strategy that organizes its procedural memory?

Prior work has progressed from extracting reusable insights across trajectories
to building evolving libraries of procedural skills from successful and
failed interactions
\citep{zhao2024expel,mi2026skillpro,zhang2026memskill,xia2026skillrl}. Recent
methods couple skill generation or curation with hierarchical, joint, or
cross-task RL \citep{li2026arise,shi2026skill1,yao2026skillrise}.
However, two challenges remain. First, existing retrieval and selection
methods typically operate directly on individual skills, leaving the
high-level strategy that governs their selection and use neither explicit nor
learned.
Second, terminal rewards reflect
the joint effects of strategy selection, skill retrieval, and execution, and
therefore cannot isolate the contribution of strategy selection.

To this end, we propose TalentRL, which separates strategy selection,
skill retrieval, and downstream execution through a predefined
hierarchical structure for procedural memory.
Coarse-grained \emph{talents} represent recurring strategies for skill use,
while fine-grained
\emph{skills} encode executable procedures. For each task, the policy
model selects a talent, which guides execution and softly prioritizes its
associated skills during retrieval. Specifically, the retriever combines task--skill similarity with
this talent prior and supplies the top-ranked skills to the policy model
during execution. Before RL, we construct talent-aware supervised fine-tuning
(SFT) demonstrations from paired rollouts to initialize both talent
routing and downstream execution.

TalentRL addresses the routing credit-assignment challenge by training
talent routing with reward contrasts across matched intervention rollouts
and optimizing downstream execution with group-relative advantages. Specifically, the interventions hold the task fixed while
withholding retrieved skills, reducing the retrieval budget, or removing
both talent guidance and its retrieval prior. These reward contrasts therefore
measure the utility of the talent-conditioned context, rather than the effect
of the talent label alone. TalentRL applies the
intervention-derived routing signal to tokens in the talent-selection
response, while tokens in subsequent execution responses receive
group-relative advantages.
Our contributions are threefold:
\begin{itemize}
    \item We propose an outcome-guided talent-aware SFT data generation pipeline
    that compares candidate talents through paired rollouts and converts
    positive-utility talent-conditioned trajectories into routing and execution
    supervision.

    \item We introduce TalentRL, a strategy-guided RL framework that combines
    hierarchical skill routing with intervention-based credit for talent
    selection and group-relative optimization of downstream execution.

    \item We demonstrate that talent-aware SFT improves
    performance across seven open-domain and multi-hop question-answering
    benchmarks, and subsequent RL achieves the highest average success rate
        among all compared methods on all three backbones.         On ALFWorld and WebShop, TalentRL achieves the
            highest ALFWorld success rate and WebShop score among all compared methods.
\end{itemize}

%% file: src/related_work.tex
\section{Related Work}

\begin{figure*}[t]
\centering
\includegraphics[width=0.98\textwidth]{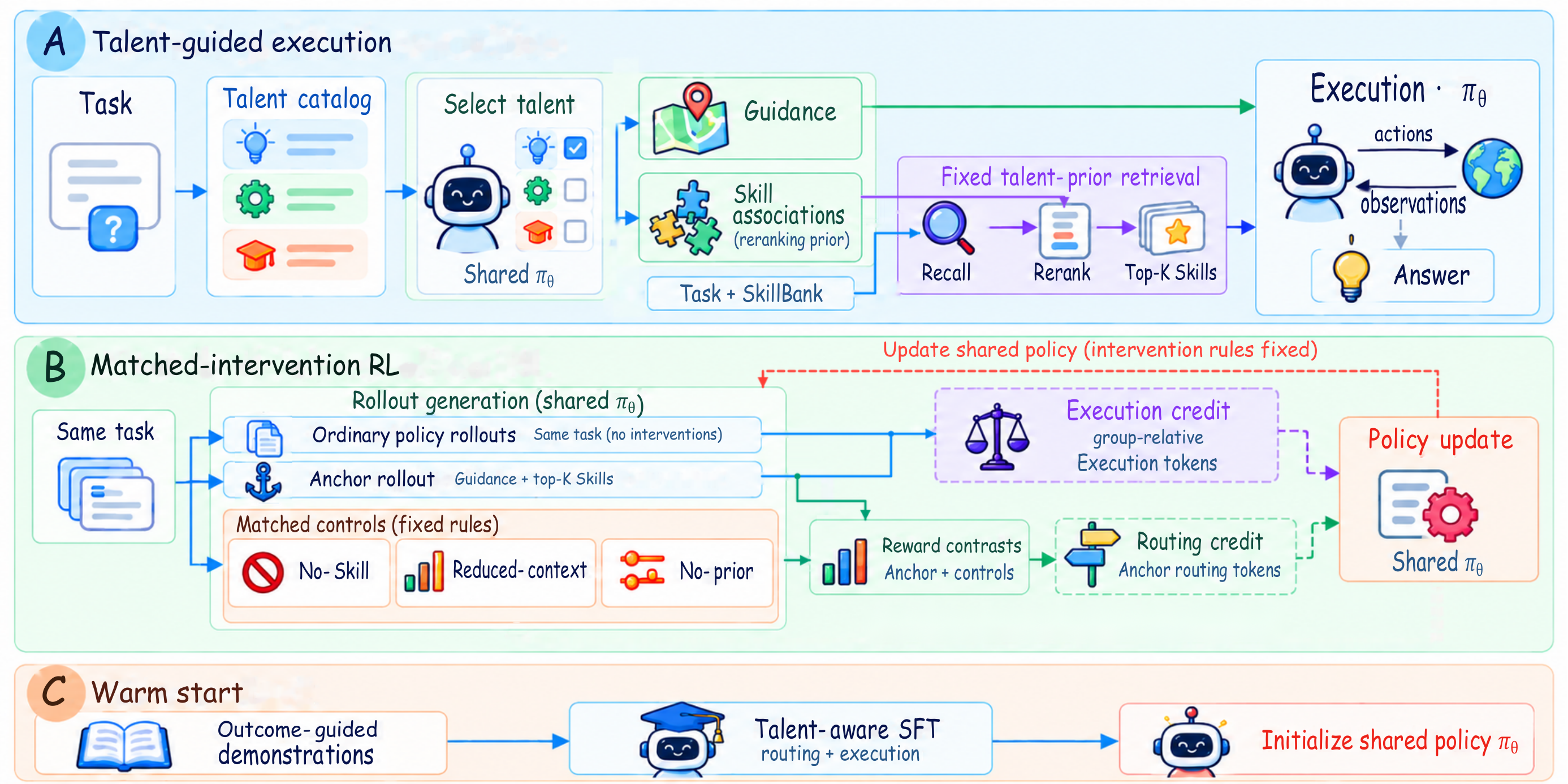}
\caption{Overview of TalentRL.
\textbf{(A)} The policy model selects a talent that guides downstream execution and provides a
soft prior for similarity-based skill retrieval.
\textbf{(B)} Reward contrasts across matched intervention rollouts train talent
routing, while group-relative advantages optimize downstream execution, both
updating the shared policy model.
\textbf{(C)} Talent-aware SFT initializes talent routing and downstream
execution from outcome-guided demonstrations.}
\label{fig:overview}
\end{figure*}

\subsection{Skill Memory and Representation}

Procedural memory, which the CoALA taxonomy \citep{sumers2024coala}
distinguishes from episodic and semantic memory, enables LLM agents to
transfer experience across tasks by storing reusable knowledge of how to act.
Existing methods represent such knowledge in different
forms and at different levels of abstraction. Natural-language representations
include verbal feedback in Reflexion \citep{shinn2023reflexion},
cross-trajectory insights in ExpeL \citep{zhao2024expel}, context-conditioned
guidelines in AutoGuide \citep{fu2024autoguide}, and step-level instructions
or higher-level scripts in MemP \citep{fang2025memp}. Buffer of Thoughts
\citep{yang2024buffer} retains reusable reasoning templates, while AWM
\citep{wang2025awm} abstracts trajectories into workflows.

Executable procedural memory represents reusable behavior as programs or
callable tools, as in Voyager \citep{wang2023voyager}, LATM
\citep{cai2024latm}, and CRAFT \citep{yuan2024craft}. Together with
natural-language memory, these approaches maintain procedural knowledge
outside model parameters, where it can be inspected, retrieved, invoked,
or composed at inference time. In contrast, SKILL0 \citep{lu2026skill0}
internalizes procedural knowledge into model parameters, while SoftSkill
\citep{tao2026softskill} encodes it as continuous representations. Our work
focuses on external natural-language skills, which remain inspectable and can
be retrieved and composed at inference time.

\subsection{Self-Evolving Skill Learning}

Self-evolving skill memory methods use accumulated agent experience to update
skill libraries. One line of work updates external skills without changing
the policy model's parameters, through procedure revision in Skill-Pro
\citep{mi2026skillpro}, validation-guided editing in SkillOpt
\citep{yang2026skillopt}, consensus-gated evolution followed by structure
specialization in SkillSpec \citep{chen2026skillspec}, or trajectory-based
consolidation in Trace2Skill \citep{ni2026trace2skill}.

Another line of work co-adapts procedural memory and the policy model. SkillRL
\citep{xia2026skillrl}, CoSkill \citep{feng2026coskill}, and SAGE
\citep{wang2025sage} couple skill-library evolution with RL. Skill1
\citep{shi2026skill1} unifies skill selection, utilization, and distillation;
SkillRise \citep{yao2026skillrise} credits skill edits according to their
cross-task utility; ReSkill \citep{he2026reskill} embeds skill creation into
GRPO-based policy optimization; SkillForge \citep{ge2026skillforge} manages
the skill library through fitness-driven skill lifecycles; and SearchSkill \citep{hu2026searchskill} evolves skills
for search. Following SkillRL \citep{xia2026skillrl}, TalentRL also evolves
the SkillBank during RL.

\subsection{Skill Routing and Credit Assignment}

Agents must determine which skills to retrieve and how to use or combine them
for a given task. One line of work develops
structured mechanisms for skill retrieval and composition. SkillRouter
\citep{zheng2026skillrouter} studies full-text retrieval and reranking at
registry scale, while SkillLens \citep{miao2026skilllens} organizes strategies,
procedures, and primitives in a multi-granularity graph. SkillComposer
\citep{zhao2026skillcomposer} jointly predicts a skill subset, its cardinality,
and the execution order of the selected skills; HyperSkill
\citep{xu2026hyperskill} connects subtasks and reusable skills through a
hypergraph.

Another line of work studies learning signals for skill selection and the
evaluation of skill use. ARISE
\citep{li2026arise} introduces a hierarchical manager--worker policy with
skill-level rewards, while RSIPO \citep{zhang2026rsipo} uses a
skill-conditioned teacher to provide token-level supervision during agentic
RL. SkillCoach
\citep{zhu2026skillcoach} assesses skill selection, following, composition, and
reflection, providing a more detailed view of skill use than final task
success alone. When the policy model selects a high-level strategy to guide
skill retrieval, terminal rewards reflect both that selection and downstream
execution, making the routing decision difficult to evaluate on its own.
TalentRL addresses this credit-assignment challenge by training talent routing
with reward contrasts across matched intervention rollouts and optimizing
downstream execution with group-relative advantages.

%% file: src/method.tex
\section{TalentRL: Strategy-Guided Skill Routing}
\label{sec:method}

Existing skill-augmented LLM frameworks such as SkillRL \citep{xia2026skillrl}
retrieve task-relevant skills by similarity before execution, without
requiring the policy model to explicitly select a high-level strategy.
In contrast, TalentRL introduces a
policy-selected talent that guides both skill retrieval and downstream
execution. We first formalize skill-augmented task execution, then describe
talent organization, routing, retrieval, and execution, followed by
talent-aware SFT data construction.

\subsection{Preliminaries}

\paragraph{LLM agents with skill memory.}
Let $\mathcal{D}$ denote the task distribution and $x\sim\mathcal{D}$ a task
description. The agent has access to a SkillBank $\mathcal{S}$ of
natural-language skills, each describing a reusable procedure. Given a task $x$, the retriever ranks
skills $s\in\mathcal{S}$ by their embedding similarity $q(x,s)$ to the task
and retrieves the top-$K$ skills:
\begin{equation}
    \mathcal{S}_{K}(x)
    =\operatorname{TopK}_{s\in\mathcal{S}} q(x,s),
\end{equation}
where $K$ is the retrieval budget and $\operatorname{TopK}$ returns the $K$
highest-scoring skills.
Let $h_0=x$. At execution turn $j\geq1$, the agent observes the history
$h_{j-1}=(x,a_1,o_1,\ldots,a_{j-1},o_{j-1})$, where $a_j$ denotes the action
at turn $j$ and $o_j$ the corresponding environment observation.
Conditioned on the interaction
history and retrieved skills, the policy model $\pi_\theta$,
parameterized by $\theta$, samples the next action:
\begin{equation}
    a_j\sim
    \pi_\theta(\cdot\mid h_{j-1},\mathcal{S}_{K}(x)).
\end{equation}
The environment returns $o_j$, and the interaction history is updated by
appending $(a_j,o_j)$ to $h_{j-1}$. Let
$\tau=(x,a_1,o_1,\ldots,a_T,o_T)$ denote the interaction trajectory, where $T$
is the number of execution turns. We maximize the expected episodic return:
\begin{equation}
\label{eq:agent-objective}
    J(\theta)
    =
    \mathbb{E}_{x\sim\mathcal{D},\,
    \tau\sim\pi_\theta(\cdot\mid x)}
    \left[R(\tau)\right],
\end{equation}
where $R(\tau)$ is the episodic return of trajectory $\tau$.
In this formulation, retrieval depends only on task similarity and is not
learned, while the policy model
learns to execute tasks using the retrieved skills. We next introduce
talent selection as a policy action that guides both retrieval and execution.

\subsection{Talent organization, routing, and execution}

Let $\mathcal{T}$ denote a catalog of talents, each representing a
coarse-grained strategy. Each talent
$c\in\mathcal{T}$ has a description $d_c$ for talent routing, guidance
$g_c$ for task execution, and a set of associated skills
$\mathcal{M}(c)\subseteq\mathcal{S}$. Each skill is associated with one or
more talents.
Following the trajectory-distillation strategy of SkillRL
\citep{xia2026skillrl}, we use an LLM to analyze successful and failed
trajectories, distill reusable procedures into skills, and identify
recurring strategies to represent as talents.
During RL, the SkillBank, talents, and talent--skill associations are updated
following SkillRL \citep{xia2026skillrl}.
Table~\ref{tab:preset-talent-skills} in Appendix~\ref{app:taxonomy} lists the
eight initial talents for search tasks and their preset skill associations.

\paragraph{Talent routing.}
TalentRL uses a shared policy $\pi_\theta$ for both routing and execution.
Before interacting with the environment, the policy model observes the task and
talent catalog and samples
\begin{equation}
    c\sim\pi_\theta(\cdot\mid x,\mathcal{T}),
    \qquad c\in\mathcal{T}.
\end{equation}
Its first response contains a routing rationale in \texttt{<think>} and the
selected talent identifier in \texttt{<talent>}; together, these fields
constitute the routing action. The following simplified example shows the policy model's
talent-selection response and the guidance supplied for execution.
\begin{routingresponse}{Talent-selection response}
\small
\textbf{Task:} \texttt{Where was the author of the novel that inspired Blade Runner born?}\\[0.35em]
\textbf{Model Response:}\\[0.1em]
\routingthinktag{think}\texttt{Identify the source novel and its author before retrieving the author's birthplace.}\routingthinktag{/think}\\
\routingtalenttag{talent}\texttt{talent\_bridge\_entity}\routingtalenttag{/talent}\\[0.5em]
\textbf{Talent Guidance ($g_c$):}\\[0.1em]
Resolve the bridge entity first, preserve relation direction, then issue a
targeted follow-up query for the requested attribute.
\end{routingresponse}

\paragraph{Talent-conditioned skill retrieval.}
For task $x$, the retriever first constructs a candidate pool
$\mathcal{R}(x)\subseteq\mathcal{S}$ of the skills with the highest similarity
$q(x,s)$. Given the
selected talent $c$, it reranks these candidates using
\begin{equation}
    \widetilde q(x,s\mid c)
    =
    q(x,s)
    +\lambda\mathbb{I}\!\left[s\in\mathcal{M}(c)\right],
    \qquad s\in\mathcal{R}(x),
\end{equation}
where $\lambda>0$ is the score bonus assigned to skills associated with $c$,
and $\mathbb{I}$ is the indicator function. The retriever then selects the
top-$K$ skills for execution:
\begin{equation}
\label{eq:talent-skill-retrieval}
    \mathcal{S}_{K}(x,c)
    =
    \operatorname{TopK}_{s\in\mathcal{R}(x)}
    \widetilde q(x,s\mid c).
\end{equation}
Talent--skill associations thus provide a soft prior rather than a
hard filter, allowing highly relevant skills outside $\mathcal{M}(c)$ to be
selected from the candidate pool. The policy model guides skill selection
through its talent choice, while the retriever selects individual
skills by reranked similarity within the budget $K$.

\paragraph{Talent-conditioned execution.}
The same policy model generates execution actions conditioned on the history, talent guidance,
and retrieved skills:
\begin{equation}
    a_j\sim
    \pi_\theta\!\left(
        \cdot\mid h_{j-1},g_c,\mathcal{S}_{K}(x,c)
    \right).
\end{equation}
\mbox{$\tau=(x,c,\mathcal{S}_{K}(x,c),a_1,o_1,\ldots,a_T,o_T)$}
denotes the resulting trajectory. The objective
in Equation~\ref{eq:agent-objective} averages over both talent selection and
task execution.

\begin{algorithm}[!t]
\caption{TalentRL}
\label{alg:talentrl}
\footnotesize
\algrenewcommand{\algorithmicrequire}{\textbf{Input:}}
\algrenewcommand{\algorithmicensure}{\textbf{Output:}}
\algtext*{EndFor}
\algtext*{EndIf}
\begin{algorithmic}[1]
\Require \begin{tabular}[t]{@{}l@{}}
Initial policy model $\pi_0$; environment $\mathcal{E}$; task sets
$\mathcal{D}_{\mathrm{sft}},\mathcal{D}_{\mathrm{rl}}$;\\
talent catalog $\mathcal{T}$; SkillBank
$\mathcal{S}$; skill budget $K$
\end{tabular}
\Ensure Trained policy $\pi_\theta$
\Statex \textcolor{cvprblue}{\textbf{// Phase I: Talent-Aware SFT}}
\State $\mathcal{D}_{\mathrm{talent}}\gets\varnothing$
\For{each task $x\in\mathcal{D}_{\mathrm{sft}}$}
    \State Collect $n$ reference rollouts $\tau_{\varnothing,r}$ with similarity-only skills.
    \State For each $c\in\mathcal{T}$, collect $n$ rollouts $\tau_{c,r}$ using $(g_c,\mathcal{S}_K(x,c))$.
    \State $\overline{u}_c(x)\gets\frac{1}{n}\sum_{r=1}^{n}u_{c,r}(x),\quad c\in\mathcal{T}$ \; (Section~\ref{sec:sft})
    \State $c^*\gets\arg\max_{c\in\mathcal{T}}\overline{u}_c(x)$
    \If{$\overline{u}_{c^*}(x)>0$}
        \State Retain a completed, valid $\tau_{c^*,r}$ with $u_{c^*,r}(x)>0$.
        \State Add its routing and execution targets to $\mathcal{D}_{\mathrm{talent}}$.
    \EndIf
\EndFor
\State Fine-tune $\pi_0$ on $\mathcal{D}_{\mathrm{talent}}$ using $\mathcal{L}_{\mathrm{sft}}$ (Eq.~\ref{eq:sft-loss}) to obtain $\pi_\theta$.
\State Initialize and freeze $\pi_{\mathrm{ref}}\gets\pi_\theta$.
\Statex \textcolor{cvprblue}{\textbf{// Phase II: Matched-Intervention RL}}
\For{each optimization iteration}
    \For{each task $x$ in a batch sampled from $\mathcal{D}_{\mathrm{rl}}$}
        \State Sample $c_{\mathrm{a}}\sim\pi_\theta(\cdot\mid x,\mathcal{T})$; collect $\tau_{\mathrm{a}}$ using $(g_{c_{\mathrm{a}}},\mathcal{S}_K(x,c_{\mathrm{a}}))$.
        \State Collect controls $\tau_{-s},\tau_k,\tau_{\mathrm{np}}$ as in Section~\ref{sec:rl-controls}.
        \State Collect ordinary rollouts; let $\mathcal{P}$ index $\tau_{\mathrm{a}}$ and the ordinary rollouts.
        \State Compute all $R_i$; obtain $\hat A_i^{\mathrm{exec}}$ for $i\in\mathcal{P}$ (Eq.~\ref{eq:execution-advantage}).
        \State Compute $\hat A^{\mathrm{talent}}$ (Eq.~\ref{eq:routing-advantage}) and assign $A_{i,t}$ (Eq.~\ref{eq:token-advantage}).
    \EndFor
    \State Update $\pi_\theta$ using the masked, KL-regularized $\mathcal{L}_{\mathrm{rl}}$ (Eq.~\ref{eq:rl-loss}).
\EndFor
\State \Return $\pi_\theta$
\end{algorithmic}
\end{algorithm}

\input{src/search_sft_table}
\subsection{Talent-aware SFT data construction}
\label{sec:sft}

Before RL, we use talent-aware SFT to initialize the policy model for both
talent routing and downstream execution. For each task $x$ in the training set
$\mathcal{D}_{\mathrm{sft}}$ and each talent $c\in\mathcal{T}$, we collect $n$
trajectories $\tau_{c,r}$ conditioned on guidance $g_c$ and retrieved skills
$\mathcal{S}_K(x,c)$, where $r=1,\ldots,n$ indexes the rollouts.
For comparison, we collect $n$ reference trajectories
$\tau_{\varnothing,r}$ for the same task using similarity-based skill
retrieval, without talent guidance or a talent-specific retrieval prior.
We define the utility of each talent-conditioned rollout as
\begin{equation}
    u_{c,r}(x)
    =
    R(\tau_{c,r})-R(\tau_{\varnothing,r}).
\end{equation}
To account for variability across rollouts, we evaluate each task--talent
pair using its mean utility,
$\overline{u}_c(x)=\frac{1}{n}\sum_{r=1}^{n}u_{c,r}(x)$.
This mean utility reflects the joint effects of talent guidance and the
retrieval prior.
We select the talent with the highest mean utility,
$c^*=\arg\max_{c\in\mathcal{T}}\overline{u}_c(x)$, and retain the task for
talent-aware data construction only if $\overline{u}_{c^*}(x)>0$.
For the selected talent $c^*$, we retain one rollout that completes normally,
contains only valid actions, and has positive utility, preferring higher
utility and fewer turns.

Each retained rollout provides supervision for both talent routing and
downstream execution. The routing target contains a short rationale in
\texttt{<think>} and the identifier of $c^*$ in
\texttt{<talent>}, and the execution targets are the actions of the
retained rollout. Individual skill selection requires no
separate supervision since the same talent-conditioned retriever handles it
during data collection, training, and inference.
Starting from the initial policy model $\pi_0$, we minimize the cross-entropy loss over
response tokens:
\begin{equation}
\label{eq:sft-loss}
    \mathcal{L}_{\mathrm{sft}}
    =
    -\sum_{t\in\mathcal{I}_{\mathrm{asst}}}
    \log\pi_\theta(y_t\mid \xi,y_{<t}),
\end{equation}
where $\xi$ denotes the input context for the current response, $y_t$ is the
target token at position $t$, $y_{<t}$ denotes the preceding tokens, and
$\mathcal{I}_{\mathrm{asst}}$ is the set of token indices in the model's
response. The resulting policy model initializes the
matched-intervention RL stage described in Section~\ref{sec:rl}.

%% file: src/search_sft_table.tex
\begin{table*}[t]
\setlength{\belowcaptionskip}{6pt}
\caption{Comparison of SkillRL-SFT and TalentRL-SFT on Search7, with zero-shot
results as a reference. Success rates (\%) are reported, with the best result
for each metric within each backbone in bold.}
\label{tab:main}
\centering
\small
\setlength{\tabcolsep}{3pt}
\renewcommand{\arraystretch}{1.08}
\begin{tabular*}{\textwidth}{@{\extracolsep{\fill}}cl|cccc|ccccc|c@{}}
\toprule
\multirow{2}{*}{\textbf{Model}} & \multirow{2}{*}{\textbf{Method}}
& \multicolumn{4}{c|}{\textbf{Single-hop}} & \multicolumn{5}{c|}{\textbf{Multi-hop}}
& \multirow{2}{*}{\textbf{Overall}} \\
& & \textbf{NQ} & \textbf{TriviaQA} & \textbf{PopQA} & \textbf{Avg.}
& \textbf{HotpotQA} & \textbf{2Wiki} & \textbf{MuSiQue} & \textbf{Bamboogle} & \textbf{Avg.} & \\
\midrule
\multirow{3}{*}{Qwen2.5-7B}
& Zero-shot & 25.32 & 50.19 & 25.07 & 33.53 & 27.63 & 25.13 & 9.76 & 27.20 & 22.43 & 27.19 \\
& SkillRL-SFT & 34.07 & 52.13 & 32.87 & 39.69 & 32.91 & 23.71 & 12.62 & 32.80 & 25.51 & 31.59 \\
& \textbf{TalentRL-SFT} & \textbf{36.29} & \textbf{53.67} & \textbf{34.26} & \textbf{41.41} & \textbf{35.30} & \textbf{27.32} & \textbf{14.60} & \textbf{41.60} & \textbf{29.70} & \textbf{34.72} \\
\midrule
\multirow{3}{*}{Qwen3-8B}
& Zero-shot & 26.12 & 54.72 & 28.51 & 36.45 & 26.17 & 29.73 & 6.70 & 35.20 & 24.45 & 29.60 \\
& SkillRL-SFT & 37.23 & 55.03 & 34.05 & 42.10 & 35.84 & 27.70 & \textbf{15.02} & \textbf{40.00} & 29.64 & 34.98 \\
& \textbf{TalentRL-SFT} & \textbf{39.39} & \textbf{56.52} & \textbf{35.52} & \textbf{43.81} & \textbf{36.35} & \textbf{33.06} & 14.11 & 39.20 & \textbf{30.68} & \textbf{36.31} \\
\midrule
\multirow{3}{*}{Llama-3.1-8B}
& Zero-shot & 21.11 & 49.64 & 27.10 & 32.62 & 21.78 & 17.41 & 6.04 & 33.60 & 19.71 & 25.24 \\
& SkillRL-SFT & 35.04 & 53.30 & 31.32 & 39.89 & 31.83 & 20.90 & 12.00 & 31.20 & 23.98 & 30.80 \\
& \textbf{TalentRL-SFT} & \textbf{38.03} & \textbf{54.60} & \textbf{35.42} & \textbf{42.68} & \textbf{32.67} & \textbf{24.40} & \textbf{12.04} & \textbf{44.00} & \textbf{28.28} & \textbf{34.45} \\
\bottomrule
\end{tabular*}
\end{table*}

%% file: src/reinforcement_learning.tex
\section{Talent-Aware RL}
\label{sec:rl}

Since an episodic return reflects both talent routing and subsequent task
execution, using it as a shared training signal does not distinguish
their respective contributions. To address this issue, TalentRL compares
matched rollouts of the same task under different routing contexts and
assigns separate credit to routing and execution tokens.

\subsection{Matched rollouts and execution credit}
\label{sec:rl-controls}

For each task, we generate an anchor rollout $\tau_{\mathrm{a}}$ using the
policy-selected talent $c_{\mathrm{a}}$, its guidance $g_{c_{\mathrm{a}}}$,
and the full skill budget $K$. We construct three matched controls.
(a) The \textbf{no-skill control} $\tau_{-s}$ retains the same talent and
guidance but withholds all retrieved skills.
(b) The \textbf{reduced-context control} $\tau_k$, where $0<k<K$, retains
the same talent, guidance, and retrieval ordering but supplies only the
top-$k$ skills.
(c) The \textbf{no-prior control} $\tau_{\mathrm{np}}$ removes both talent
guidance and the talent-specific retrieval bonus while retaining the
full budget of similarity-retrieved skills.
We collect these matched rollouts
alongside ordinary, independently sampled policy rollouts for the same
task.
Let $R_i=R(\tau_i)$ denote the return of rollout $i$.
Let $\mathcal{P}$ denote the index set of the anchor and ordinary policy
rollouts for the same task, excluding the matched control rollouts. Following GRPO
\citep{shao2024deepseekmath}, the execution advantage for each
$i\in\mathcal{P}$ is
\begin{equation}
\label{eq:execution-advantage}
 \hat A_i^{\mathrm{exec}}
 =
 \frac{R_i-\mu_{\mathcal{P}}}
 {\sigma_{\mathcal{P}}+\epsilon}.
\end{equation}
Here, $\mu_{\mathcal{P}}$ and $\sigma_{\mathcal{P}}$ are the mean and standard
deviation of returns in $\mathcal{P}$, and $\epsilon>0$ is a numerical
stability constant. Routing credit is computed by comparing the anchor
return with the control returns, as described next.

\subsection{Matched-intervention routing credit}

Let $R_{\mathrm{a}}$, $R_{-s}$, $R_k$, and $R_{\mathrm{np}}$ denote the
task returns of the anchor, no-skill, reduced-context, and no-prior
rollouts, respectively. To ignore return differences with magnitude below a
threshold $\eta\geq0$, we define a dead-zone operator:
\begin{equation}
D_\eta(z)=
\begin{cases}
0,& |z|<\eta,\\
z,& \text{otherwise}
\end{cases}.
\end{equation}
Let $\mathcal{J}=\{\mathrm{skill},\mathrm{budget},\mathrm{talent}\}$ index
the three reward contrasts. We compute
\begin{equation}
\delta_\ell=D_\eta\!\bigl(R_{\mathrm{a}}-R_\ell\bigr),
\quad \ell\in\mathcal{J}.
\end{equation}
Here, $R_\ell$ is the return of the no-skill, reduced-context,
or no-prior control for $\ell=\mathrm{skill},\mathrm{budget},\mathrm{talent}$,
respectively.
We combine these contrasts into a routing advantage:
\begin{equation}
\label{eq:routing-advantage}
\hat A^{\mathrm{talent}}
=\operatorname{clip}\!\left(
    \alpha\sum_{\ell\in\mathcal{J}}w_\ell\delta_\ell,\,-\kappa,\,\kappa
\right),
\end{equation}
where $w_\ell\geq0$ and $\sum_{\ell\in\mathcal{J}}w_\ell=1$.
The weights control the relative contributions of the three contrasts,
while $\alpha>0$ scales the combined signal and $\kappa>0$ sets its clipping
bound. This routing advantage provides task-specific feedback on the utility of the
talent-conditioned context.

\input{src/search_results_tables}
\subsection{Token-level policy objective}

We optimize the shared policy model by assigning routing and execution
advantages to their corresponding response tokens. Let $m_{i,t}=1$ if token
$t$ in rollout $i$ is generated by the policy model, either during routing
or execution, and $m_{i,t}=0$ for input tokens, including the task prompt,
talent guidance, retrieved skills, and environment observations. This mask
excludes input tokens from the policy surrogate loss; $A_{i,t}$ determines
the advantage assigned to each generated token.
Let $\mathcal{I}_i^{\mathrm{route}}$ and $\mathcal{I}_i^{\mathrm{exec}}$
denote the routing and execution token positions in rollout $i$,
respectively. Let $i_{\mathrm{a}}$ denote the index of the anchor rollout.
We define the token-specific advantage as
\begin{equation}
\label{eq:token-advantage}
A_{i,t}=
\begin{cases}
\hat A^{\mathrm{talent}},
& i=i_{\mathrm{a}},\ t\in\mathcal{I}_i^{\mathrm{route}},\\[3pt]
\hat A_i^{\mathrm{exec}},
& i\in\mathcal{P},\ t\in\mathcal{I}_i^{\mathrm{exec}},\\[3pt]
0,& \text{otherwise}.
\end{cases}
\end{equation}
Let $r_{i,t}(\theta)$ denote the probability ratio for the sampled token at
position $t$ in rollout $i$ under the current policy and the behavior policy
that generated the rollout, conditioned on the same context.
We minimize the following KL-regularized clipped surrogate objective:
\begin{equation}
\label{eq:rl-loss}
\begin{aligned}
&\mathcal{L}_{\mathrm{rl}}(\theta)
=-\frac{1}{N}\sum_{i,t}m_{i,t}\,\ell_{i,t}(\theta)
+\beta D_{\mathrm{KL}}\!\left(\pi_\theta\Vert\pi_{\mathrm{ref}}\right),\\
&\ell_{i,t}(\theta)
=\min\!\bigl(r_{i,t}(\theta)A_{i,t},
\operatorname{clip}\!\left(r_{i,t}(\theta),1-\epsilon_c,1+\epsilon_c\right)A_{i,t}\bigr).
\end{aligned}
\end{equation}
Here, $N=\sum_{i,t}m_{i,t}$ counts the response tokens, $\epsilon_c$ controls
probability-ratio clipping, and $\beta$ controls the strength of KL
regularization toward the frozen reference policy $\pi_{\mathrm{ref}}$,
initialized from the SFT checkpoint.
The routing rationale and talent identifier receive the same
routing advantage since together they form the routing action.
Algorithm~\ref{alg:talentrl} summarizes the complete training procedure,
from talent-aware SFT to matched-intervention RL.

%% file: src/search_results_tables.tex
\begin{table*}[t]
\setlength{\belowcaptionskip}{6pt}
\caption{Search7 success rate (\%) after RL.
Within each backbone, the highest and second-highest distinct values for each
metric are bold and underlined, respectively; ties receive the same formatting.
Evaluation details are provided in
Appendix~\ref{app:experimental-configuration}.}
\label{tab:rl-summary}
\centering
\small
\setlength{\tabcolsep}{3pt}
\renewcommand{\arraystretch}{1.08}
\begin{tabular*}{\textwidth}{@{\extracolsep{\fill}}cl|cccc|ccccc|c@{}}
\toprule
\multirow{2}{*}{\textbf{Model}} & \multirow{2}{*}{\textbf{Method}}
& \multicolumn{4}{c|}{\textbf{Single-hop}} & \multicolumn{5}{c|}{\textbf{Multi-hop}}
& \multirow{2}{*}{\textbf{Overall}} \\
& & \textbf{NQ} & \textbf{TriviaQA} & \textbf{PopQA} & \textbf{Avg.}
& \textbf{HotpotQA} & \textbf{2Wiki} & \textbf{MuSiQue} & \textbf{Bamboogle} & \textbf{Avg.} & \\
\midrule
\multirow{6}{*}{Qwen2.5-7B}
& CoSkill & 44.24 & 61.39 & 38.09 & 47.91 & 37.87 & 39.09 & 14.81 & 39.20 & 32.74 & 39.24 \\
& SAGE & \underline{46.43} & \textbf{63.60} & 39.30 & 49.78 & 41.67 & \underline{41.57} & 16.92 & 42.40 & \underline{35.64} & 41.70 \\
& Skill1 & 46.15 & 62.66 & 38.70 & 49.17 & 39.41 & 39.44 & 15.52 & 38.40 & 33.19 & 40.04 \\
& ReSkill & 46.20 & 62.96 & 39.97 & 49.71 & \underline{43.08} & 37.64 & 17.00 & 42.40 & 35.03 & 41.32 \\
& SkillRL & \textbf{48.01} & 63.24 & \textbf{40.70} & \textbf{50.65} & 43.04 & 36.90 & \underline{17.38} & \underline{44.00} & 35.33 & \underline{41.90} \\
& \textbf{TalentRL} & 45.93 & \underline{63.30} & \underline{40.68} & \underline{49.97} & \textbf{43.24} & \textbf{42.16} & \textbf{17.83} & \textbf{48.80} & \textbf{38.01} & \textbf{43.13} \\
\midrule
\multirow{6}{*}{Qwen3-8B}
& CoSkill & 44.32 & 62.66 & 39.30 & 48.76 & 39.55 & 41.12 & 15.76 & 42.40 & 34.71 & 40.73 \\
& SAGE & 45.62 & 61.96 & \textbf{41.32} & 49.63 & 41.39 & \textbf{44.18} & 16.76 & 44.00 & 36.58 & 42.18 \\
& Skill1 & 46.68 & 62.62 & 40.86 & 50.05 & 43.05 & 41.81 & 18.54 & \underline{45.60} & \underline{37.25} & 42.74 \\
& ReSkill & 46.48 & 62.69 & 40.09 & 49.75 & \underline{44.15} & 39.38 & \textbf{19.03} & 44.00 & 36.64 & 42.26 \\
& SkillRL & \underline{47.53} & \textbf{64.16} & \underline{41.24} & \underline{50.98} & 43.98 & 39.03 & 18.08 & \underline{45.60} & 36.67 & \underline{42.80} \\
& \textbf{TalentRL} & \textbf{48.28} & \underline{64.00} & 41.09 & \textbf{51.12} & \textbf{45.00} & \underline{43.05} & \underline{18.62} & \textbf{47.20} & \textbf{38.47} & \textbf{43.89} \\
\midrule
\multirow{6}{*}{Llama-3.1-8B}
& CoSkill & 45.76 & 63.94 & 40.21 & 49.97 & 38.14 & 37.75 & 15.43 & \underline{44.00} & 33.83 & 40.75 \\
& SAGE & \textbf{49.78} & 64.79 & \underline{42.29} & \underline{52.29} & \textbf{43.44} & \textbf{42.68} & 17.13 & 41.60 & 36.21 & \underline{43.10} \\
& Skill1 & 48.17 & 64.78 & 41.38 & 51.44 & 42.31 & \underline{42.60} & \textbf{17.38} & 43.20 & \underline{36.37} & 42.83 \\
& ReSkill & \underline{49.42} & \underline{65.15} & 41.94 & 52.17 & \underline{43.30} & 39.13 & 15.72 & \underline{44.00} & 35.54 & 42.66 \\
& SkillRL & 47.98 & 63.12 & 41.35 & 50.82 & 42.43 & 41.89 & 17.05 & 40.00 & 35.34 & 41.97 \\
& \textbf{TalentRL} & 48.75 & \textbf{66.03} & \textbf{44.09} & \textbf{52.96} & 42.43 & 39.09 & \underline{17.17} & \textbf{48.80} & \textbf{36.87} & \textbf{43.77} \\
\bottomrule
\end{tabular*}
\end{table*}

%% file: src/experiment.tex
\section{Experiments}
\label{sec:experiments}

\begin{table*}[t]
\setlength{\belowcaptionskip}{6pt}
\caption{ALFWorld success rate and WebShop average score/success rate (\%).
Baseline values are paper-reported, not protocol-matched reproductions.
Plain SFT and TalentRL-SFT are Qwen2.5-7B-Instruct students; their ALFWorld
columns use the 140 valid-seen games. TalentRL is initialized from the
TalentRL-SFT checkpoints.
Best and second-best results are bold and underlined, respectively.}
\label{tab:interactive-results}
\centering
\small
\setlength{\tabcolsep}{3.5pt}
\renewcommand{\arraystretch}{1.12}
\begin{tabular*}{0.96\textwidth}{@{\extracolsep{\fill}}lrrrrrrrrr@{}}
\toprule
\textbf{Method} & \multicolumn{7}{c}{\textbf{ALFWorld Success Rate (\%)}}
& \multicolumn{2}{c}{\textbf{WebShop}} \\
\cmidrule(lr){2-8}\cmidrule(lr){9-10}
& \textbf{Pick} & \textbf{Look} & \textbf{Clean} & \textbf{Heat}
& \textbf{Cool} & \textbf{Pick2} & \textbf{Overall}
& \textbf{Score} & \textbf{Succ.} \\
\midrule
MemRL          & 62.8 & 38.5 & 22.2 & 12.5 & 8.0 & 0.0
               & 21.4 & 29.5 & 9.2 \\
EvolveR        & 64.9 & 33.3 & 46.4 & 13.3 & 33.3 & 33.3
               & 43.8 & 42.5 & 17.6 \\
Mem0+GRPO      & 78.1 & 54.8 & 56.1 & 31.0 & 65.0 & 26.9
               & 54.7 & 58.1 & 37.5 \\
SimpleMem+GRPO & 89.5 & 36.3 & 60.0 & 50.0 & 64.9 & 26.3
               & 62.5 & 67.8 & 46.9 \\
SkillRL        & \underline{97.9} & 71.4 & 90.0
               & 90.0 & 95.5 & 87.5
               & 89.9 & 85.2 & 72.7 \\
Skill1         & \textbf{100.0} & \underline{98.6} & \underline{97.3}
               & \underline{99.2} & \textbf{96.1} & \underline{96.0}
               & \underline{97.5} & \underline{89.7} & \textbf{82.9} \\
\midrule
Plain SFT      & \textbf{100.0} & \textbf{100.0} & 74.1 & \textbf{100.0}
               & 84.0 & 91.7 & 90.7 & 73.0 & 58.4 \\
TalentRL-SFT   & \textbf{100.0} & \textbf{100.0} & 81.5
               & \textbf{100.0} & 88.0 & 95.8
               & 93.6 & 75.8 & 63.0 \\
\textbf{TalentRL} & \textbf{100.0} & \textbf{100.0} & \textbf{100.0}
                 & \textbf{100.0} & \underline{96.0} & \textbf{100.0}
                 & \textbf{99.3} & \textbf{90.7} & \underline{79.6} \\
\bottomrule
\end{tabular*}
\end{table*}

\subsection{General setup}

We use Qwen2.5-7B-Instruct \citep{yang2024qwen25}, Qwen3-8B
\citep{yang2025qwen3}, and Llama-3.1-8B-Instruct
\citep{grattafiori2024llama3} as backbone models.
We evaluate on seven open-domain and multi-hop question-answering benchmarks:
Natural Questions \citep{kwiatkowski2019nq}, TriviaQA
\citep{joshi2017triviaqa}, PopQA \citep{mallen2023popqa}, HotpotQA
\citep{yang2018hotpotqa}, 2WikiMultiHopQA \citep{ho2020twowiki}, MuSiQue
\citep{trivedi2022musique}, and Bamboogle
\citep{press2023compositionality}. We refer to this suite as Search7 and
report the average success rate (\%) across the seven benchmarks, with equal
weight assigned to each benchmark.
We also evaluate on two interactive benchmarks,
ALFWorld \citep{shridhar2021alfworld} for household tasks and WebShop
\citep{yao2022webshop} for online shopping.
We use GRPO for RL training with a batch size of 64 and 8 rollouts per task.
RL runs for 200 steps on the NQ and HotpotQA training sets for Search7 and
150 steps on ALFWorld and WebShop, with five skills retrieved by
Qwen3-Embedding-0.6B \citep{zhang2025qwen3embedding} and $\lambda = 0.3$.
Further experimental details are provided in
Appendix~\ref{app:experimental-configuration}.

\subsection{Baselines}
\label{sec:baselines}

We consider the following baselines that use RL to improve the policy model
with an external skill library. (a)~\textbf{CoSkill} \citep{feng2026coskill}
jointly trains reasoning and meta-skill agents to evolve a hierarchical SkillBank.
(b)~\textbf{SkillRL} \citep{xia2026skillrl} retrieves procedural skills and
recursively evolves its skill library during RL.
(c)~\textbf{SAGE} \citep{wang2025sage} couples RL with a self-improving external skill library.
(d)~\textbf{Skill1} \citep{shi2026skill1} unifies skill
selection, utilization, and distillation in one policy.
(e)~\textbf{ReSkill} \citep{he2026reskill} embeds skill creation into
GRPO-based policy optimization.
On ALFWorld and WebShop, we additionally compare with the memory-based agents
MemRL \citep{zhang2026memrl}, EvolveR \citep{wu2025evolver}, Mem0
\citep{chhikara2025mem0}, and SimpleMem \citep{liu2026simplemem}, using the results reported by
SkillRL \citep{xia2026skillrl}.

\subsection{Main results}
\label{sec:main-results}

\paragraph{Talent-aware SFT strengthens policy initialization.}
To examine whether talent-aware supervision improves policy initialization,
we start from the search SFT data released by SkillRL \citep{xia2026skillrl},
label its tasks with talents following Section~\ref{sec:sft}, and add the
resulting talent-conditioned trajectories to the original SFT trajectories.
Starting from each zero-shot backbone, we separately perform full-parameter
fine-tuning on this augmented data (TalentRL-SFT) and on the original
trajectories alone (SkillRL-SFT), and compare the two in Table~\ref{tab:main}.
TalentRL-SFT reaches average Search7 success rates of 34.72\%, 36.31\%, and
34.45\% on Qwen2.5-7B, Qwen3-8B, and Llama-3.1-8B, respectively, improving
over SkillRL-SFT by +3.13\%, +1.33\%, and +3.65\%.
TalentRL-SFT improves over SkillRL-SFT on both the single-hop and multi-hop
averages for every backbone, by +1.72\%, +1.71\%, and +2.79\% on single-hop
benchmarks and by +4.19\%, +1.04\%, and +4.30\% on multi-hop benchmarks for
Qwen2.5-7B, Qwen3-8B, and Llama-3.1-8B, respectively.
The multi-hop gains are driven mainly by Bamboogle on Qwen2.5-7B
(+8.80\%) and Llama-3.1-8B (+12.80\%) and by 2Wiki on Qwen3-8B (+5.36\%),
whereas Qwen3-8B trails slightly on MuSiQue (\(-0.91\)\%) and Bamboogle
(\(-0.80\)\%), which explains its smaller multi-hop gain.
Across all three model architectures, the combined routing and execution
supervision provides a stronger policy initialization for subsequent RL.

\begin{figure*}[t]
\centering
\includegraphics[width=\textwidth]{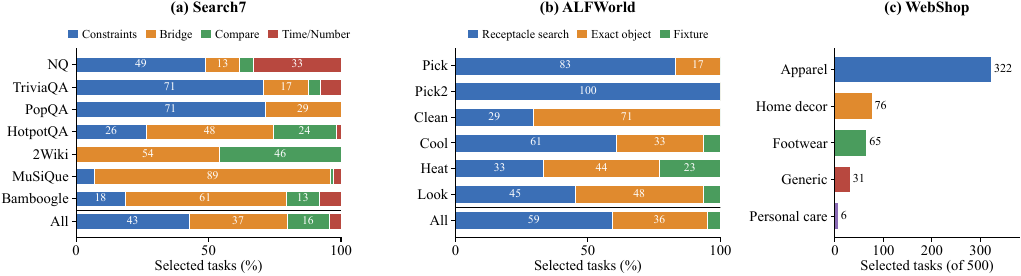}
\caption{Talent selection of the final TalentRL policies. (a) Search7 test
sets, as \% of questions per dataset. (b) ALFWorld by task type over the
seen (140) and unseen (134) splits. (c) WebShop on 500 goals. Percentages
below 10\% are omitted from the bars.}
\label{fig:talent-distribution}
\end{figure*}

\paragraph{TalentRL on Search7 Benchmarks.}
We then examine whether talent-aware RL further improves the policy model.
Starting from each TalentRL-SFT checkpoint, we train the policy model with
the objective in Section~\ref{sec:rl}, whereas all other baselines in
Table~\ref{tab:rl-summary} start from the corresponding SkillRL-SFT
checkpoint. We compare TalentRL with these baselines in
Table~\ref{tab:rl-summary} and with the published interactive-task results
in Table~\ref{tab:interactive-results}.
As shown in Table~\ref{tab:rl-summary}, TalentRL reaches average Search7
success rates of 43.13\%, 43.89\%, and 43.77\% on Qwen2.5-7B, Qwen3-8B, and
Llama-3.1-8B, respectively, achieving the highest average among all compared
methods on each backbone. Relative to the strongest baseline on each
backbone, TalentRL improves the average success rate by +1.23\% over SkillRL
on Qwen2.5-7B, +1.09\% over SkillRL on Qwen3-8B, and +0.67\% over SAGE on
Llama-3.1-8B.
On multi-hop benchmarks, TalentRL achieves the highest average on all three
backbones, exceeding the strongest baseline by +2.37\% over SAGE on
Qwen2.5-7B, +1.22\% over Skill1 on Qwen3-8B, and +0.50\% over Skill1 on
Llama-3.1-8B.
The advantage is most pronounced on Bamboogle, where TalentRL ranks first on
all three backbones and exceeds the second-best method by 4.80\%, 1.60\%, and
4.80\%, respectively.
On single-hop benchmarks, it achieves the highest average on Qwen3-8B and
Llama-3.1-8B and the second highest on Qwen2.5-7B, where it trails SkillRL by
0.68\%.

\paragraph{Interactive-task results.}
Table~\ref{tab:interactive-results} presents ALFWorld and WebShop results.
We use GPT-5.5 to generate SFT trajectories with retrieved skills. Plain SFT
fine-tunes Qwen2.5-7B-Instruct on the successful trajectories, and
TalentRL-SFT further fine-tunes this checkpoint on talent-labeled
trajectories constructed as in Section~\ref{sec:sft}
(Appendix~\ref{app:experimental-configuration}). Without interactive RL,
TalentRL-SFT raises ALFWorld overall success
from 90.7\% to 93.6\% (131/140), with the largest gains on Clean
(+7.4\%), Pick2 (+4.2\%), and Cool (+4.0\%). On WebShop, it raises the
average score from 73.0 to 75.8 and the success rate from 58.4\% to 63.0\%
(315/500).
Interactive RL from the TalentRL-SFT checkpoints further improves both
benchmarks. On ALFWorld, TalentRL reaches 99.3\% overall success
(139/140), +5.7\% over TalentRL-SFT and the best among all methods,
with 100\% on every task type except Cool (96.0\%).
On WebShop, TalentRL improves over TalentRL-SFT by +14.9 in average score and
+16.6\% in success rate, reaching the best average score among all methods
(90.7) and a success rate of 79.6\% (398/500), second only to Skill1
(82.9\%).

\paragraph{Learned talent routing.}
Figure~\ref{fig:talent-distribution} shows that the trained policies select
different talents for different types of tasks, and the selected talent
matches the structure of each task. On Search7,
most TriviaQA and PopQA questions ask for a single attribute of a single entity and
are routed mainly to \talentname{constraint application}, which narrows candidate answers
by the conditions stated in the question. NQ is also mostly single-hop, but
many of its questions ask \emph{when} or \emph{how many}, and 33\% of them
are routed to \talentname{temporal and numeric reasoning}, compared with at most 8\% on
the other benchmarks. MuSiQue and Bamboogle require resolving an intermediate
entity first and are routed mainly to \talentname{bridge-entity resolution} (89\% and
61\%). HotpotQA and 2Wiki also contain comparison questions such as
\emph{Who was born first, A or B?}, and 24\% and 46\% of their questions are
routed to \talentname{evidence comparison}. On ALFWorld, the selected talent reflects the main bottleneck
of each task type. Pick and Pick2 tasks, which are limited by locating
objects, are routed mainly to \talentname{systematic receptacle search}, Clean tasks mainly
to \talentname{exact object selection}, and Heat tasks, which require operating the
microwave, select \talentname{fixture interaction} more often than any other task type (23\%). On WebShop, the
talents correspond to product domains, and the selected talent follows the
category of the requested product. All 322 goals
routed to \talentname{apparel} request clothing. The 31 goals routed to \talentname{generic matching} fall outside the four specific domains and are mainly electronics and food.
Appendix~\ref{app:talent-distribution} gives further details.

%% file: src/conclusion.tex
\section{Conclusion}

We propose TalentRL, a reinforcement learning framework in which a single policy model learns both to select a talent for each task and to execute skills under that talent. The selected talent guides how the policy model solves the task and adds a soft prior to skill retrieval.
Talent-aware SFT teaches the policy model this select-then-execute format, and RL then trains talent selection with matched-intervention contrasts and skill execution with group-relative advantages.
On Search7, TalentRL improves over prior skill-based methods after both
SFT and RL on all three backbones, with the largest gains on multi-hop
questions, and it remains competitive on ALFWorld and WebShop. Future work will extend TalentRL to more domains and larger models, and explore discovering new talents directly from RL experience.

%% file: src/skill_card_details.tex
\begingroup
\small
\captionsetup[longtable]{singlelinecheck=false}
\setlength{\tabcolsep}{4pt}
\setlength{\defaultaddspace}{0.6em}
\setlength{\LTleft}{0pt}
\setlength{\LTright}{0pt}
\setlength{\LTcapwidth}{\textwidth}
\renewcommand{\arraystretch}{1.2}
\begin{longtable}{@{}
>{\raggedright\arraybackslash}p{0.14\textwidth}
>{\raggedright\arraybackslash}p{0.50\textwidth}
>{\raggedright\arraybackslash}p{\dimexpr0.36\textwidth-4\tabcolsep\relax}@{}}
\caption{Details of the preset SkillBank skills referenced by the Search7
talent catalog. Principles and applicability conditions reproduce the
corresponding skill fields; only typography is normalized.}
\label{tab:skill-card-details}\\[6pt]
\toprule
\textbf{Skill ID} & \textbf{Principle} & \textbf{When to apply} \\
\midrule
\endfirsthead
\multicolumn{3}{@{}l}{\small\itshape\tablename~\thetable\ continued from the previous page}\\[6pt]
\toprule
\textbf{Skill ID} & \textbf{Principle} & \textbf{When to apply} \\
\midrule
\endhead
\midrule
\multicolumn{3}{r@{}}{\footnotesize\itshape Continued on the next page}\\
\endfoot
\bottomrule
\endlastfoot
\texttt{gen\_001}
& Break the question into minimal sub-questions (entities, attributes, relations) and handle each with its own query before synthesizing the final answer.
& Any complex or multi-part question, especially multi-hop or comparative tasks. \\
\addlinespace
\texttt{gen\_003}
& If top results don't yield direct evidence, rephrase by adding qualifiers (dates, roles, context) or alternate names instead of repeating the same query.
& After scanning first result set and finding no definitive evidence. \\
\addlinespace
\texttt{gen\_004}
& Only state an answer once the supporting evidence (reliable source snippet) has been located and read; refrain from guessing or summarizing from memory.
& Before committing to a final answer, especially for contentious or less-known facts. \\
\addlinespace
\texttt{gen\_005}
& Validate critical facts (dates, numbers, names) against at least two independent sources to avoid outdated or incorrect information.
& When the fact may have changed over time or when initial sources disagree. \\
\addlinespace
\texttt{gen\_006}
& Detect ambiguous entity mentions and disambiguate by adding clarifiers (e.g., occupation, year, nationality) in the query or by exploring disambiguation pages.
& When an entity name maps to multiple possible referents or results appear mixed. \\
\addlinespace
\texttt{gen\_007}
& For multi-hop tasks, first retrieve the intermediate entity (e.g., a film's director) then run a second targeted query for that entity's attribute (e.g., birthplace) before synthesizing.
& Whenever the answer depends on an attribute of a related entity rather than the entity in the question. \\
\addlinespace
\texttt{gen\_008}
& When comparing entities (earlier/later, larger/smaller), collect each relevant value separately, list them side by side, then decide using explicit comparison logic.
& Any comparative question involving dates, counts, or quantitative attributes. \\
\addlinespace
\texttt{gen\_009}
& For time-sensitive queries (e.g., upcoming releases, current versions), include temporal cues like ``latest'', year, or ``release date 2023'' and prioritize most recent authoritative sources.
& Questions asking for ``current'', ``next'', or ``latest'' information. \\
\addlinespace
\texttt{gen\_010}
& Stop issuing further queries once you have clear, corroborated evidence; conversely, avoid premature termination if no source explicitly supports an answer.
& After each read step---decide to answer only if confidence is justified; otherwise refine search. \\
\addlinespace
\texttt{dir\_002}
& If first search yields weak/no hits, instantly reformulate using synonyms, alternate names, dates, or quoted phrases instead of repeating the same query.
& After an initial search returns no clear answer or only tangential results. \\
\addlinespace
\texttt{dir\_003}
& For song titles, quotes, episode names, etc., wrap the unique phrase in quotation marks to pull exact-match sources.
& When the query contains distinctive phrases, lyrics, book/film titles, or direct quotations. \\
\addlinespace
\texttt{dir\_004}
& Include recency cues (e.g., ``current'', year) in the search and verify publication date to avoid outdated or speculative info.
& For questions about `current', `latest', or future events/releases. \\
\addlinespace
\texttt{dir\_005}
& Only state an answer that is explicitly supported by the retrieved text; if unclear, continue searching rather than guess or hallucinate.
& Before finalizing any factoid answer. \\
\addlinespace
\texttt{mul\_001}
& Split the main query into explicit sub-questions to identify each entity and the specific attribute or relationship needed before searching.
& Any multi-hop question that links two or more entities/facts (e.g., director's birthplace, album vs film comparisons). \\
\addlinespace
\texttt{mul\_002}
& Issue separate, focused searches for each sub-question or entity instead of one broad query to gather precise intermediate facts.
& After decomposition, when distinct pieces of information must be collected individually. \\
\addlinespace
\texttt{mul\_003}
& Retrieve concrete values for all items involved (dates, places, relations) before performing any comparison or conclusion to avoid premature or unsupported answers.
& Comparative tasks (earlier/later, older/younger, bigger/smaller, etc.). \\
\addlinespace
\texttt{mul\_004}
& Add clarifying descriptors (profession, year, medium, location) to queries to distinguish between entities with identical or similar names and prevent mixing facts.
& Whenever an entity name is generic or shared by multiple subjects (e.g., common personal names, titles appearing in different media). \\
\addlinespace
\texttt{mul\_005}
& If initial search fails or returns ambiguous results, promptly rephrase using synonyms, alternate titles, or additional context instead of repeating the same query.
& After a search yields no relevant results or conflicting information. \\
\addlinespace
\texttt{ent\_002}
& If the name is common or ambiguous, append clarifiers (birth year, nationality, known work) to the query or scan result snippets for matching context before selecting a source.
& When multiple people/characters share the same name or early results mix different entities. \\
\addlinespace
\texttt{ent\_003}
& Confirm the attribute in at least two independent, reputable sources (e.g., Wikipedia infobox + biography site) to avoid hallucinations.
& After the first plausible answer appears or when the attribute seems uncommon or uncertain. \\
\addlinespace
\texttt{ent\_004}
& If initial search returns irrelevant hits, adjust the query instead of repeating it---use alternative spellings, middle initials, or related works/titles.
& When consecutive top results don't mention the sought attribute for the intended entity. \\
\addlinespace
\texttt{com\_001}
& First split the question into (a) each entity and (b) the single attribute to be compared; this ensures every subsequent search is targeted and comparable.
& At the moment you read any comparison-type question. \\
\addlinespace
\texttt{com\_002}
& Independently retrieve the identical attribute for each entity---run separate, attribute-focused searches (e.g., ``[Entity] headquarters'', ``[Entity] release date'') and store the raw values.
& Immediately after identifying entities and the comparison attribute. \\
\addlinespace
\texttt{com\_003}
& Convert retrieved values to a common, directly comparable form (e.g., standardize dates, compute ages, map demonyms to countries) before judging equality or ordering.
& After gathering each entity's attribute but before drawing any conclusion. \\
\addlinespace
\texttt{com\_004}
& If a search yields no clear or multiple results, refine the query with context (profession, year, language) or acknowledge data insufficiency---never infer without evidence.
& Whenever an attribute lookup returns ambiguous, missing, or conflicting information. \\
\addlinespace
\texttt{com\_005}
& State each normalized value and perform the logical comparison (equal, earlier, larger, etc.) explicitly to derive the answer, ensuring traceability and avoiding hallucination.
& As the final synthesis step before outputting the comparison result. \\
\end{longtable}
\endgroup